\documentclass[10pt,twocolumn,letterpaper]{article}

\usepackage{wacv}
\usepackage{times}
\usepackage{epsfig}
\usepackage{graphicx}
\usepackage{amsmath}
\usepackage{amssymb}

\usepackage[pagebackref=true,breaklinks=true,letterpaper=true,colorlinks,bookmarks=false]{hyperref}

\wacvfinalcopy % *** Uncomment this line for the final submission

\def\wacvPaperID{248} % *** Enter the wacv Paper ID here

\ifwacvfinal\fi
\begin{document}

%%%%%%%%% TITLE
\title{Actor Conditioned Attention Maps for Video Action Detection}

\author{Oytun Ulutan$^1$ \hspace{1.0cm} Swati Rallapalli$^2$ \hspace{1.0cm} Mudhakar Srivatsa$^2$ \\
Carlos Torres$^1$\hspace{1.0cm} B.S. Manjunath$^1$\\ \\
$^1$University of California, Santa Barbara  \hspace{1.0cm} $^2$IBM T. J. Watson Research Centre\\
\vspace{-1.0cm}
% \and
}

\maketitle
\ifwacvfinal\thispagestyle{empty}\fi

%%%%%%%%% ABSTRACT
\begin{abstract}
   While observing complex events with multiple actors, humans do not assess each actor separately, but infer from the context. The surrounding context provides essential information for understanding actions. To this end, we propose to replace region of interest(RoI) pooling with an attention module, which ranks each spatio-temporal region's relevance to a detected actor instead of cropping. We refer to these as Actor-Conditioned Attention Maps (ACAM), which amplify/dampen the features extracted from the entire scene. The resulting actor-conditioned features focus the model on regions that are relevant to the conditioned actor. For actor localization, we leverage pre-trained object detectors, which transfer better. The proposed model is efficient and our action detection pipeline achieves near real-time performance. Experimental results on AVA 2.1 and JHMDB demonstrate the effectiveness of attention maps, with improvements of $7$ mAP on AVA and $4$ mAP on JHMDB. 
\end{abstract}
%%%%%%%%% BODY TEXT
\vspace{-0.5cm}
\section{Introduction}
% \vspace*{-.2cm}
\noindent \textbf{Motivation:} Human action detection is a promising field, which can improve applications such as surveillance, robotics and autonomous driving. While many datasets (e.g., HMDB-51\cite{Kuehne11}, Kinetics\cite{kay2017kinetics}, UCF-101\cite{soomro2012ucf101}) are very useful for video search and classification, a recent AVA\cite{gu2018ava} dataset focuses on atomic actions within short video segments. Atomic actions have the potential to transfer to different contexts, become building blocks for more complex actions and improve the general understanding of human actions/interactions in videos. For these reasons, in this work we focus on the task of atomic action detection from videos. We propose to model actor actions by using information from the surrounding context and evaluate our model on AVA\cite{gu2018ava} and JHMDB\cite{Jhuang:ICCV:2013} datasets. We demonstrate the efficiency and transferability of our approach by implementing an action detection pipeline and qualitatively testing it on videos from various sources.

\noindent \textbf{Challenges:}
While observing actions/activities, humans infer from the entire context and our perception depends on the surrounding objects, actors, and scene. This is a concept that has been widely studied in neuroscience and psychology \cite{eckstein2013rethinking, henderson2008full, preston2013neural, torralba2006contextual}. The idea of explicitly leveraging context is directly relevant to our action detection task as surroundings of actors provide valuable information. 

\begin{figure}[t]
\begin{center}
\includegraphics[width=1.0\linewidth]{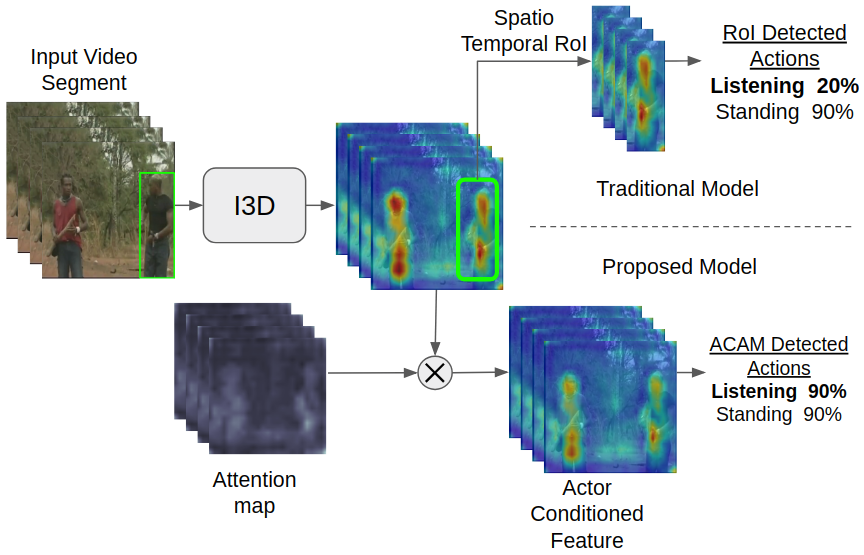}
\end{center}
\vspace*{-.25cm}

   \caption{Comparing RoI pooling with the proposed ACAM method for video action detection. ACAM explicitly models the surrounding context and generates features from the complete scene by conditioning them on detected actors. For example, presence of a talking person next to the actor is evidence for the ``listening" action, which is captured by attention maps. 
    \vspace{-0.5cm}
   }
\label{fig:acam_vs_roi}
\end{figure}

Studies in action detection task have followed the ideas from the R-CNN architectures and extended it to videos\cite{gu2018ava, peng2016multi, saha2016deep, singh2017online}. However, in action detection, the bounding box locates the \textit{actor} rather than the \textit{action} itself and datasets do not include explicit interaction labels for the actions, which makes it challenging to model context. In such setting, RoIPooling becomes insufficient for modeling actions and including the contextual information such as actors, objects and scene. In order to address this, we propose attention maps as a replacement for RoIPooling for action detection. The proposed methodology learns context in a weakly supervised manner as demonstrated in Fig. \ref{fig:acam_vs_roi}.

\noindent \textbf{Approach:}
% Contextual modeling has been used in recent works. 
% Non-local Neural Networks \cite{wang2018non} model the contextual information by generating a weighted sum of global features at every feature location and compressing the surrounding scene context. Research in Visual Question Answering  \cite{das2017human, lu2016hierarchical, yang2016stacked} uses attention maps to model relevant interactions in an image and focus the model to answer questions. Actor-Centric Relations Network \cite{sun2018actor} modeled relations of actors by generating contextual features for each detected actor.
Inspired by the context \cite{wang2018non}, attention \cite{vaswani2017attention} and relation \cite{sun2018actor} ideas, the proposed model generates attention maps \textit{conditioned on each actor from contextual and actor features}. Replacing the traditional way of cropping an actor RoI from the context feature maps, generated attention maps multiply and scale the context feature maps according to each actor. The scaling operation amplifies relevant regions and dampens the irrelevant regions to the conditioned actor. This allows the model to learn the complex interactions for the current actor with the scene which may include surrounding actors, objects and background.
% are generated for each feature dimension and determine the relation of actor to every spatio-temporal context location. Such an attention mechanism allows us to focus on the actor without cropping as in RoIPooling while capturing the spatio-temporal structure of the scene.

%-------------------------------------------------------------------------
%\subsection{Contributions}
\vspace*{.2cm}
\noindent \textbf{Technical Contributions}
% \footnote{Code will be made available at \href{}{Github}. A real-time demo of the pipeline in Section \ref{sec:pipeline} is available at \href{}{Repo}}: 
\begin{itemize}
\vspace*{-.25cm}
    \item \textbf{Generation of ACAMs:} We propose an attention mechanism for person action detection that models the surrounding context of actors. These maps replace the RoIPooling with attention maps, which amplify and dampen regions conditioned on the actor.
    % These maps condition the contextual features on the actors in a weakly supervised way without explicit interaction labels.
\vspace*{-.25cm}
    \item \textbf{Object detectors as transferable and modular region proposal networks (RPN): }Instead of retraining an RPN on the dataset, we use a pre-trained object detector to obtain accurate actor locations and demonstrate that it is more transferable to unseen data. 
\vspace*{-.25cm}
    \item \textbf{End-to-end pipeline for real-time video action detection on videos: } We implement a pipeline and qualitatively show that our approach transfers well on videos from various types of unseen sources. 
\end{itemize}

% Codes will be made available at \href{https://github.com/oulutan/ActorConditionedAttentionMaps}{Github}. A real-time demo is also available at \href{https://github.com/oulutan/ACAM_Demo}{Demo Repo}.
Code will be made available at \href{https://github.com/oulutan/ActorConditionedAttentionMaps}{Github}. A real-time demo of the pipeline from Section \ref{sec:pipeline} is also available at \href{https://github.com/oulutan/ACAM_Demo}{Github}. %[LINKS REMOVED FOR ANONYMITY]

%-------------------------------------------------------------------------
\section{Related Work}
State of the art models on the earlier action recognition datasets \cite{Kuehne11, sigurdsson2016hollywood, soomro2012ucf101}  use models such as Two-Stream networks~\cite{simonyan2014two} combining RGB with Optical Flow, 2D Convolutions with LSTMs \cite{yue2015beyond} and 3D Convolutions \cite{hou2017tube}. The release of the large-scale, high quality datasets like Sports 1M \cite{KarpathyCVPR14}, Kinetics \cite{kay2017kinetics}, allowed deeper 3D CNN models such as C3D~\cite{tran2015learning}, Inception 3D  (I3D)~\cite{carreira2017quo} to be trained and achieve high performance. 
% Recent work focuses on temporal action detection from untrimmed videos (e.g.,  ActivityNet \cite{caba2015activitynet}, THUMOS \cite{idrees2017thumos}) using two-Stream 2D CNNs \cite{zhao2017temporal}, LSTMs~\cite{yeung2018every} and 3D CNNs~\cite{shou2016temporal}.% are used in this task.

In this work, we are focusing on the Atomic Visual Actions (AVA v2.1) \cite{gu2018ava} dataset. This dataset exhaustively annotates the atomic actions and spatial locations of all the actors. Initial methods on the AVA dataset extended the Faster-RCNN \cite{ren2015faster} architectures to 3D convolutions, where initial layers generate actor proposals and each proposal is analyzed by subsequent layers \cite{gu2018ava}. The recent Actor Centric Relation Network (ACRN) \cite{sun2018actor} model generates features by combining actor and scenes to represent actor's interactions with surrounding context. Our proposed model leverages this relation idea to generate attention maps.

% Attention models are used in Natural Language Processing (NLP) \cite{bahdanau2014neural,rush2015neural, vaswani2017attention}. 
An attention function for relating different positions of a sentence was used in \cite{vaswani2017attention}. 
% Studies in Visual Question Answering task focused on generating attention maps from the input question to focus visual model \cite{das2017human, lu2016hierarchical, yang2016stacked}. 
The relation module in \cite{santoro2017simple} combines questions with vision to generate answers. \cite{hu2018relation} uses object relations to effectively detect them. 
% This approach improves both instance recognition and duplicate removal. 
An LSTM structure is used in \cite{li2017attentive} to generate an attention map to model contextual information. Inspired by these attention models which use positional relations, our proposed model generates a spatio-temporal attention map which scales different positions of the feature map depending on each actor. 

Recently, \cite{wang2018non} used a compact feature representation that compresses non-local information from contextual features with a weighted sum of pixels for action detection and achieved state-of-the-art results. This shows that contextual information is essential and detection can be improved by replacing the RoIPooling which ignores context. 
% In a zero-shot learning setting \cite{mettes2017spatial} uses existence of objects and their locations to detect interactions. 

Context has also been studied on image action detection. V-COCO \cite{gupta2015visual} and HICO-Det \cite{chaolearning} datasets have exhaustive annotations on persons, objects and their interactions. Unlike our task where the interactions are weakly supervised, these annotations enable models to learn interactions efficiently.
Interaction modeling from \cite{gkioxari2017detecting} achieved good performance with a multi-stream network  where each stream focused on people, objects and interactions separately. 

Our proposed model builds on top of the relation idea from ACRN \cite{sun2018actor} where the relation between the actor and the surrounding context is generated. Unlike ACRN where the relation features are used for classification, our model leverages the relation function to generate attention maps. This is similar to \cite{vaswani2017attention} where attention maps localize relevant parts in a feature map. However, in our task, the attention maps condition the features to each actor in the scene individually. Since the feature maps are conditioned individually and include context, ACAM models actor-context information more effectively than RoIPooling. 

%-------------------------------------------------------------------------
% \section{Actor Conditioned Attention Maps}
\section{Proposed Method}
\begin{figure*}[t]
\begin{center}
\includegraphics[width=1.0\linewidth]{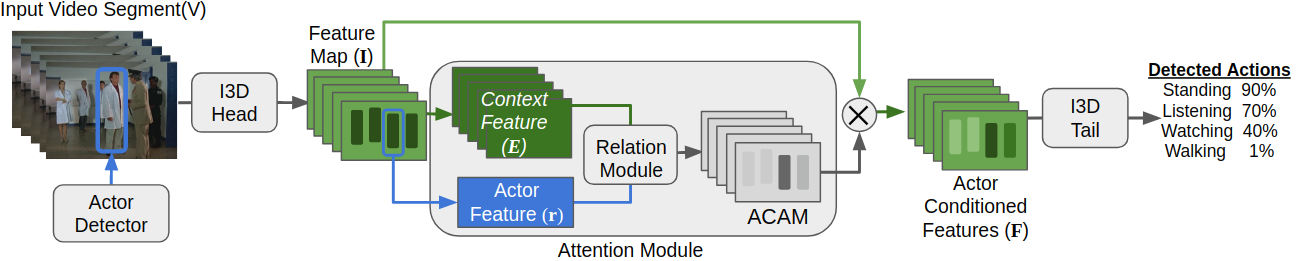}
\end{center}
\vspace*{-.25cm}
   \caption{ACAM architecture. The input video segments are processed by the I3D back-bone. Feature vectors for each detected actor are generated from their locations on the feature map. A set of weights is generated for every spatio-temporal region in the scene by combining the actor features and contextual features extracted from the entire scene. These weights, i.e., attention maps, are multiplied by the feature map and the result represents the actor conditioned features. Four detected actors are represented by four vertical bars in $I$. One focused actor (boxed) is listening to a close-by actor. This action is captured by larger weights in the attention map shown as a darker vertical bar.
   \vspace*{-0.25cm}}
\label{fig:model_arch}
\end{figure*} 

This section describes our proposed model for action detection. From each input video segment, the objective is to detect bounding boxes for each actor and classify their actions. Each actor can have multiple action labels (e.g., ``sitting" and ``talking" simultaneously). 

\subsection{Context for Atomic Actions}

Compared to object detection tasks, action boundaries are ill defined and can include interactions with the surrounding context (objects, actors and scene). Different actions require different sizes of visual areas to be considered from the input video. For example, the ``walking" action requires the model to consider only the pixels on the actor and close surrounding context, whereas the ``listening" action requires the model to look for at a larger context area (e.g., a talking person) around the actor in addition to the actor itself. With such variety in action classes, using traditional object detection methods such as RoIPooling can potentially lose the contextual information around the actors. Even though features cropped using RoIPooling include information from a larger receptive field, this technique compresses the information into a smaller feature map
% , loses the spatio-temporal ordering of the surroundings 
and does not explicitly model interactions (with other actors and context). Additionally, large-scale video datasets do not provide explicit interaction labels (person 1 is listening to person 2) but weak labels (person 1 - listening, person 2 - talking). These interactions need to be learned via weak supervision. In order to address these challenges, the proposed method generates attention maps for each detected actor to model the importance of each spatio-temporal region in the feature map by conditioning on detected actors.
The proposed model architecture is shown in Fig. \ref{fig:model_arch}.

\subsection{Actor Conditioned Attention Maps}

The proposed method generates a set of weights (ACAMs) that represent the attention of different parts of the spatio-temporal input. Our action detection problem contains multiple actors performing concurrent actions that can be either related or disparate. This generates an attention task, where different actors relate differently to spatio-temporal locations in the scene. The proposed model addresses the attention problem by generating ACAMs, which capture relations between actors and context. Instead of extending RoIPooling \cite{ren2015faster} to action detection as in \cite{gu2018ava}, the essence of ACAMs is to condition the features extracted from the entire scene on each actor and action dynamics.

Spatio-temporal features are extracted from the input video $V$ with a 3D convolutional back-bone (e.g.,: I3D~\cite{carreira2017quo}) up-to some layer (Mixed\_4f).
Let $\mathbf{I}$ represent the extracted feature tensor of size $(T\times H \times W \times C)$ with temporal resolution $T$ and spatial resolution $H\times W$ indexed by $t$, $h$, and $w$ and feature channel dimension $C$. 
i.e.,  $ \mathbf{I} = conv3d(V)$.

The actor feature vector $\mathbf{r}_a$ of size $N$ (set to $C/4$) is extracted for actor $a$ using RoI pooling extended time via: 

\begin{equation}
    \mathbf{r}_a = \phi( \mathbf{w_\rho} RoI(\mathbf{I}, a) + \mathbf{b_\rho}),
\end{equation}

\noindent where $\phi(x) = ReLU(x) = max(0,x)$, $\mathbf{w_\rho}$ are the weights, and $\mathbf{b_\rho}$ are the biases. Similar to Faster RCNN \cite{ren2015faster}, $\mathbf{r}_a$ can be used for classifying the actions of actor $a$. Instead of using $\mathbf{r}_a$ directly, we propose to leverage its descriptive potential to generate relations between the actor and the context.

The conditioned feature vector $\mathbf{F}_{t,h,w|a}$ is computed for each actor $a$ in the scene and spatio-temporal indices $(t,h,w)$. This is generated by a conditioning function of actor feature $\mathbf{r}_a$ and contextual features $\mathbf{I}_{t,h,w}$ via:

\begin{equation}
    \mathbf{F}_{t,h,w|a} = Condition(\mathbf{I}_{t,h,w} | \mathbf{r}_a,), ~\forall~ (t,h,w)
\end{equation}

Following steps explain the $Condition$ function. Activations in $\mathbf{I}$ are sparse; however, it is compressed by an additional layer to obtain a denser representation ($\mathbf{E}$) via: 
\begin{equation}
    \mathbf{E}_{t,h,w} = \phi ( \mathbf{w}_\eta\mathbf{I}_{t,h,w} + \mathbf{b}_\eta),
\label{eq:embedding}
\end{equation}

\noindent where $\mathbf{w}_\eta$ and $\mathbf{b}_\eta$ are the weights and biases. The new tensor $\mathbf{E}$ has shape $(T\times H \times W \times M)$ with $M < C$ (set $M=C/4$). This approach reduces the dimensionality of $\mathbf{I}$ and captures higher level information similar to \cite{wang2018non, carreira2017quo, ulutan2018order}.

The relation tensor for actor $a$ (i.e., $\mathbf{R}_a$) is inspired by the ``relation" idea from \cite{santoro2017simple} and it is modified to capture the relations between actor $a$ and every location $t,h,w$ in the context as: 

\begin{equation}
    \mathbf{R}_{a,t,h,w} = \mathbf{w}_\Omega \mathbf{r}_a + \mathbf{w}_\gamma \mathbf{E}_{t,h,w} + \mathbf{b}_\beta,
\label{eq:relation}
\end{equation}

\noindent where $\mathbf{w}_\Omega$ and $\mathbf{w}_\gamma$ are the weights for actor and context features, respectively; and $\mathbf{b}_\beta$ are the biases. $\mathbf{R}_{a,t,h,w}$ describe the relation of actor features and contextual locations. We set up the shapes of $\mathbf{w}_\Omega, \mathbf{w}_\gamma, \mathbf{b}_\beta$ such that $\mathbf{R_a}$ has the same shape as $\mathbf{I}$. %, which is $(T\times H \times W \times C)$. 
Note that ReLU ($\phi$) is not used in Eq. \ref{eq:relation}.

\begin{figure}[t]
\begin{center}
\includegraphics[width=0.9\linewidth]{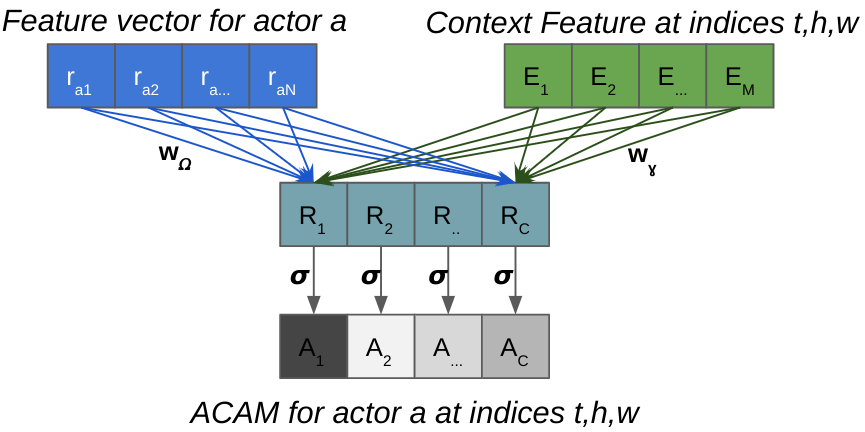}
\end{center}
 \vspace*{-.25cm}
  \caption{Attention module for actor $a$ at a single index $t,h,w$. Attention weights in ACAM at index $t,h,w$ are generated from the actor feature $\mathbf{r}_a$ and context features at the same index $\mathbf{E}_{t,h,w}$.
\vspace*{-.5cm}
  }
\label{fig:relation_module}
\end{figure}

Instead of using relation features for classification directly, we leverage the I3D back-bone and its pre-trained weights by conditioning $\textbf{I}$ on the actor $a$ for an increased performance (Section \ref{ss:modules}). Inspired by the ``forget" gates of LSTMs, attention module generates the actor conditioned attention maps for $a$ (i.e., $\mathbf{ACAM}_a$) by:

\begin{equation}
    \mathbf{ACAM}_{a,t,h,w} = \sigma(\mathbf{R}_{a,t,h,w}),
\label{eq:sigmoid}
\end{equation}

\noindent and conditioned features $\mathbf{F}$ as follows:

\begin{equation}
    \mathbf{F}_{t,h,w|a} = \mathbf{I}_{t,h,w} \odot \mathbf{ACAM}_{a,t,h,w}
\label{eq:elementwise}
\end{equation}

\noindent where $\sigma$ is the sigmoid function which scales the attention maps in $[0,1]$ interval and $\odot$ is the elementwise multiplication of vectors. 
Attention maps multiplied by $\mathbf{I}_{t,h,w}$ weights the different regions on the context. This process amplifies regions relevant to actor $a$, while damping the irrelevant regions. The generation of ACAMs is shown in Fig. \ref{fig:relation_module} for actor $a$ at a single spatio-temporal index $t,h,w$.

\begin{figure}[t]
\begin{center}
\includegraphics[width=1.0\linewidth]{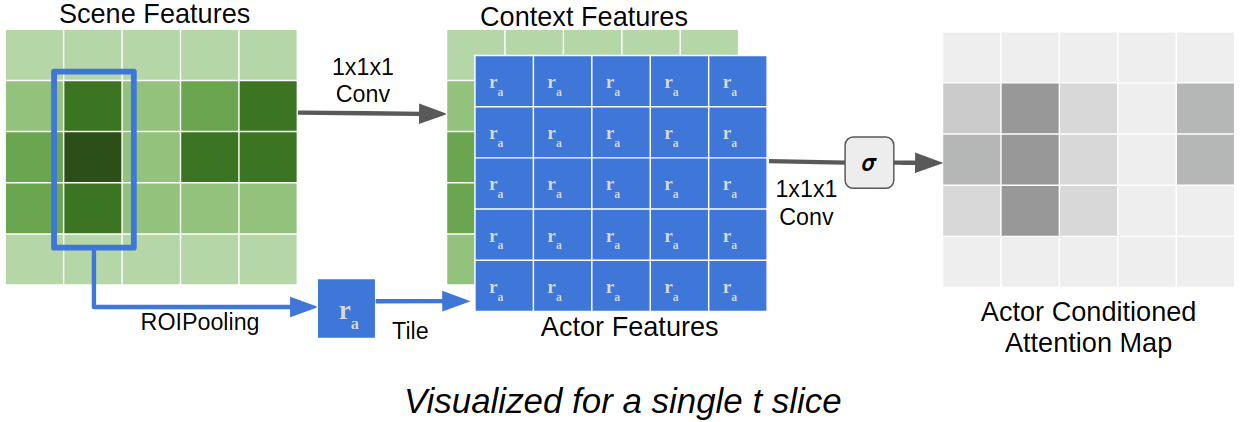}
\end{center}
\vspace*{-.25cm}
  \caption{Calculation of attention maps with convolutions. Actor feature from the RoI is tiled and concatenated to features extracted from the context at every spatio-temporal index. Convolutions on the combined feature calculate the relations from Eq \ref{eq:relation} efficiently.
  \vspace{-0.3cm}}
\label{fig:acam}
\end{figure}

These operations are efficiently computed using $1\times 1 \times 1$ convolutions. For instance, Eq. \ref{eq:embedding} are fully connected layers repeated for every $t,h,w$ index, which is equivalent to $1\times 1 \times 1$ convolutions. In Eq. \ref{eq:relation}, $\mathbf{r}_a$ is constant for all indices $t,h,w$ as shown in Fig. \ref{fig:acam}. This equation is computed by repeating $\mathbf{r}_a$ of shape $1\times 1 \times 1 \times N$ to match the spatio-temporal shape of $\mathbf{E}$. The repeated actor feature has shape $T\times H \times W \times N$ and is concatenated with $\mathbf{E}$ to produce a tensor with shape $T\times H \times W \times (N+M)$. Applying the $1\times 1 \times 1$ convolutions to the concatenated tensor is equivalent to Eq. \ref{eq:relation} and produces $\mathbf{R}$. The sigmoid operation (Eq. \ref{eq:sigmoid}) on $\mathbf{R}$ generates the attention maps ($\mathbf{ACAM}$). Element-wise multiplication from Eq. \ref{eq:elementwise} generates $\textbf{F}$, which is then classified by remaining layers of the CNN back-bone.

\subsection{Person Detectors as RPN}

We experiment with pre-trained(COCO) frozen and fine-tuned(AVA) person detectors for actor localization. Our approach has the following three advantages over RPNs:

\noindent \textbf{1. Transferability:} Object detectors see large object variations (MS-COCO \cite{lin2014microsoft}). This allows models trained on object detection datasets to transfer to videos from different sources. Action datasets, however, usually come from similar sources such as AVA (Movies), JHMDB (Youtube), which reduces the diversity in actor views and limits transferability of fine-tuned solutions for actor localization.

\noindent \textbf{2. Efficiency:} ACAM requires fewer actor proposals than RoI pooling to enable its complex computations. These detections are obtained from pre-trained person detectors.
% ACAM feature representation uses more memory compared to RoI pooling. To train efficiently, fewer number of accurate proposals are essential and these can be obtained by pre-trained object detectors.

\noindent \textbf{3. Modularity:} The action model is trained using a slow and highly accurate actor detector. The modularity of the proposed methodology enables replacing detectors based on performance and application requirements. For example, a faster detector used for testing achieves near real-time performance, which is demonstrated in Section \ref{sec:pipeline}.
% We demonstrate these advantages in Section \ref{sec:ablation}.

%-------------------------------------------------------------------------
\section{Experiments and Evaluations}
% \subsection{Dataset and Evaluation}
\subsection{Datasets and Implementation Details}
\noindent \textbf{Datasets:} The proposed ACAM model is tested on the AVA v2.1\cite{gu2018ava} and JHMDB\cite{Jhuang:ICCV:2013} datasets. 

\textbf{AVA} contains 2-second video segments of multiple actors with 211k training and 57k validation samples. Actor bounding boxes are annotated for the center frames only. Weak action labels are provided for the complete segment without temporal localization or explicit interactions. Actors can have multiple action labels in each segment. We follow the AVA v2.1 evaluation process and calculate the mean Average Precision (mAP) across 60 classes. There are three action super classes Person Poses (13 classes), Object Interactions (32 classes), Person Interactions (15 classes). 

\textbf{JHMDB} contains 1-second video segments of 21 action classes across 928 video clips with single actor-action pairs. 

% \subsection{Implementation Details}
% \textbf{Implementation Details: }
\noindent \textbf{3D CNNs:} 
We use I3D \cite{carreira2017quo} as the 3D CNN back-bone for all of our model candidates. The input video segment of RGB frames is processed by the initial I3D layers until the ``Mixed\_4f" layer to obtain the feature tensor $\mathbf{I}$ of size $8\times25\times25\times832$. The actor conditioned features $\mathbf{F}$ are computed using ACAM, where $\mathbf{F}$ is a weighted version of the original feature map ($\mathbf{I}$). The remaining I3D layers are used and initialized with pre-trained weights. We use the remaining layers up to final ``Mixed\_5c" for classification on $\mathbf{F}$ and call this operation ``I3D Tail". A global average pooling across spatio-temporal dimensions is applied to the final feature map to compute class probabilities. Each 2-second video is uniformly subsampled down to 32 frames. 

\noindent \textbf{Actor Detection:} Detectors process all the videos and store the detected actors locations. We use the Faster R-CNN \cite{ren2015faster} with NAS \cite{zoph2017learning} detector pre-trained on MS-COCO \cite{lin2014microsoft} dataset. Additionally, we fine-tune the detector for actors and compare their performances on AVA and transferability on other datasets. This object detector is further analyzed and available in Tensorflow Object Detection API \cite{huang2017speed}. 
% We change the available default configuration to allow batch processing and  reduce the detection confidence threshold to 1\% (Default 30\%) as AP metric requires to rank the low confidence detections.

\noindent \textbf{Data Augmentations:} In addition to cropping and flipping the video sequences, we augment the actor box coordinates from the detector. This generates a slight difference in extracted $\mathbf{r}_a$ at each training step and reduces overfitting.
% allows the model to be less dependent on the object detector and allows object detector to be switched with a faster detector during test time.

\noindent \textbf{Training:} We initialize our models with I3D weights trained on Kinetics-400 dataset \cite{carreira2017quo} and train our models with Adam optimizer \cite{kingma2014adam} and cosine learning rate \cite{loshchilov2016sgdr} between max (0.02) and min (0.0001) for 70 epochs. We use a batch size of 2 per GPU and 4 Nvidia 1080Ti (total batch size of 8). Batch-norm updates are disabled. All models are implemented in Tensorflow \cite{tensorflow2015-whitepaper}.

\subsection{Comparisons with the State of the Art }

\begin{table}
\begin{small}
\begin{center}
  \begin{tabular}{| l || c | }
    \hline
    Model Architecture      & AVA v2.1 Validation mAP \\ \hline \hline
    Single Frame\cite{gu2018ava}          & 14.20 \\ \hline
    I3D \cite{gu2018ava}                  & 15.10 \\ \hline
    ACRN \cite{sun2018actor}                & 17.40 \\ \hline
    ACAM & 23.29 \\ \hline
    \textbf{ACAM - tuned} & \textbf{24.38} \\ \hline
  \end{tabular}
\end{center}
\caption{Validation mAP results compared to published state of the art results. Proposed ACAM model with fine-tuned actor detector achieves the highest performance on the AVA v2.1 Validation set. 
\vspace{-0.2cm}
}
\label{tab:published_map_results}
\end{small}
\end{table}

% \textbf{State of the art results: }
% Table \ref{tab:published_map_results} shows that compared to ACRN\cite{sun2018actor}, the highest performing paper on AVA dataset, our ACAM model improves performance of validation set by $~5mAP$. 

Table \ref{tab:published_map_results} shows ACAM with fine-tuned object detector outperforming the recent ACRN \cite{sun2018actor} on AVA validation set by $~7$mAP. We compare our model with validation results of the models from ``ActivityNet 2018 AVA challenge''\cite{ghanem2018activitynet}. Table \ref{tab:challenge_map_results} shows that ACAM outperforms in validation. The table excludes results from ensemble models and focuses on comparing their highest performing single RGB model. 

Additionally, we compare the effects of fine-tuning the actor detector on AVA dataset. Comparing ACAM with \textbf{ACAM-tuned} demonstrates the improvement gained by fine-tuning the actor detector on the AVA dataset. However, this comes with trade-offs as analyzed in Section \ref{sec:ablation}.

% The Deep Mind \cite{girdhar2018better} model weights and code from the challenge are not available. Therefore, we implemented a model similar for performance comparisons and refer to it as ``I3D Head + RoIPool + I3D Tail''. This in-house implementation achieves 19.83 mAP, while the performance compared to ACAM's 23.29 demonstrates that the proposed ACAM method is complementary to architectures that use RoIPooling and can increase the performance of \cite{girdhar2018better}.

\begin{table}
\begin{small}
\begin{center}
  \begin{tabular}{| l || c | }
    \hline
    Model Architecture      & AVA v2.1 Validation mAP \\ \hline \hline
    YH Technologies\cite{yao2018yh}          & 19.40 \\ \hline
    Megvii/Tsinghua\cite{jianghuman}         & 20.01 \\ \hline
    Deep Mind\cite{girdhar2018better}        & 21.90 \\ \hline
    
    ACAM & 23.29 \\ \hline
    \textbf{ACAM - tuned} & \textbf{24.38} \\ \hline

  \end{tabular}
\end{center}
\caption{ACAM mAP results compared with models from the ActivityNet CVPR-2018 AVA challenge. We excluded the ensemble/fusion methods to evaluate the benefits of the proposed layer.
\vspace{-0.5cm}
}
\label{tab:challenge_map_results}
\end{small}
\end{table}

\subsection{Comparisons of Individual Modules} \label{ss:modules}

\begin{figure*}
\begin{center}
\includegraphics[width=0.9\linewidth]{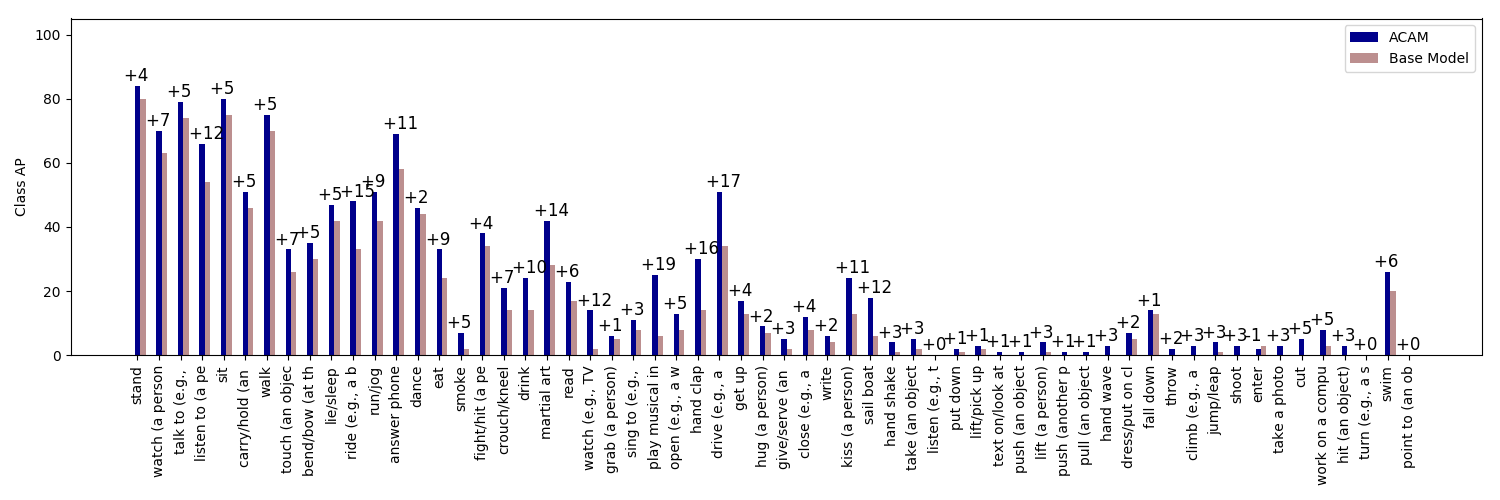}
\end{center}
\vspace*{-.25cm}
   \caption{Per class AP results for the proposed ACAM model and the base model I3D Head + RoIPool on the AVA dataset. The classes are sorted by the number of training samples available in the dataset. Improvements achieved by ACAM are visualized on the bars.
   \vspace{-0.4cm}
   }
\label{fig:per_class_ap}
\end{figure*}

% The proposed model architecture consists of multiple modules. 
In this section, we demonstrate the purpose of each module and experiment with alternative models to ACAM for representing contextual interactions. These experiments use actor detectors that are pre-trained, frozen on COCO dataset and are not fine-tuned for AVA. AVA mAP results, inference speeds and number of parameters for these implementations are shown in Table \ref{tab:baselines_map_results} to analyze the trade-offs.

\noindent \textbf{I3D Head + RoIPool (Base Model):} The base in-house implementation follows the model from \cite{gu2018ava} and achieves $18.01$ mAP. The input video goes through I3D convolutions upto the layer ``Mixed\_4f" and call it I3D Head. Using RoI pooling the actor feature vector $\mathbf{r}_a$ is obtained and used with fully connected layers for classification.

\noindent \textbf{I3D Head + RoIPool + I3D Tail:} This model uses RoI pooling to extract actor features. Instead of vectorizing the RoI and using fully connected layers, we use the remaining I3D layers from ``Mixed\_4f" to ``Mixed\_5c" similar to the model from \cite{girdhar2018better}. This method achieves $19.83$ mAP and demonstrates that using I3D Tail improves the performance. 

\begin{table}
\begin{footnotesize}
\begin{center}
  \begin{tabular}{| l || c | c | c | }
    \hline
    Model Architecture      & \begin{tabular}[c]{@{}c@{}}AVA\\mAP\end{tabular} & Test Speed & \# Params  \\ \hline \hline
    I3D Head + RoIPool (Base)  & 18.01 & 8.3 samples/s & 7,573,244 \\ \hline
    I3D Head + RoIPool + Tail  & 19.83 & 8.1 samples/s & 12,341,484 \\ \hline
    I3D Head + ACRN   + Tail   & 20.59 & 7.1 samples/s & 13,034,956 \\ \hline
    I3D Head + NL-RoI + Tail   & 20.82 & 7.5 samples/s & 13,035,164 \\ \hline
     \textbf{I3D Head + ACAM + Tail} & \textbf{23.29} & 6.9 samples/s & 13,034,956 \\ \hline
  \end{tabular}
\end{center}
\caption{mAP results of our different variants. We compare ACAM with alternate attention modules and base models. Inference speed is measured on a single 1080Ti GPU. Number of parameters include I3D Head and Tail parameters for applicable models. 
% \vspace{-0.5cm}
}
\label{tab:baselines_map_results}
\end{footnotesize}
\end{table}

\begin{table}
\begin{small}
\begin{center}
  \begin{tabular}{| l || c | c | c | }
    \hline
    Model Architecture      &  Pose  & Objects & Interaction \\ \hline \hline
    I3D Head + RoIPool        & 36.88 &  9.87 & 19.02\\ \hline
    I3D Head + RoIPool + Tail  & 38.45 & 12.11 & 20.16\\ \hline
    I3D Head + ACRN    + Tail   & 38.38 & 12.52 & 22.37 \\ \hline
    I3D Head + NL-RoI + Tail   & 40.50 & 12.06 & 22.45 \\ \hline
    \textbf{I3D Head + ACAM + Tail} & \textbf{42.13} & \textbf{15.02} & \textbf{24.22}  \\ \hline

  \end{tabular}
\end{center}
\caption{mAP comparisons of different RoI variants on AVA action super classes. Pose: Person Pose actions (ex: walking, standing), Objects: Object Manipulation (ex: drink, pull), Interaction: Person Interaction (ex: talk to a person, watch a person). 
\vspace{-0.3cm}
}
\label{tab:category_map}
\end{small}
\end{table}

\noindent \textbf{I3D Head + NL-RoI + I3D Tail:} Non-Local Neural Networks \cite{wang2018non} uses a compact representation to model interactions between different spatio-temporal regions in a video segment. We modify this model to generate non-local features between the detected actor features $\mathbf{r}_a$ and scene context features $\mathbf{I}$. This model achieves $20.82$ mAP.  

\noindent \textbf{I3D Head + ACRN + I3D Tail:} Similar to the ACRN \cite{sun2018actor}, this implementation classifies on the relation features ($\mathbf{R}$). To improve performance, we use ``I3D Tail" instead of the added $3\times3$ convolutions. This model achieves $20.59$ mAP. Comparing ACAM to this model demonstrates that attention based context is better than relation features alone. 

\noindent \textbf{I3D Head + ACAM + I3D Tail:} This model uses the proposed ACAMs to condition context features on actors and classifies the actions using I3D Tail. This model achieves $23.29$ mAP, which is the highest performance when compared to the alternative relation models. 

The breakdown of performance per action super class is demonstrated in Table \ref{tab:category_map}. 
% The super classes are defined by the AVA dataset \cite{gu2018ava} and 
the AP values in the table are averaged across super classes. This experiment demonstrates leveraging contextual information with ACAM improves the performance for every super class. 

The per class performance (AP) comparison on the proposed ACAM model and the base model is shown in Fig. \ref{fig:per_class_ap}. We observe significant (above 10 AP) improvements in passive actions such as ``listen a person" and ``watch TV" as in those classes context is active. Scene context improves the detection of classes such as ``drive" and ``play instrument".

\subsection{Results on JHMDB} \label{sec:jhmdb}

In addition to the AVA dataset, we evaluate our models on the JHMDB \cite{Jhuang:ICCV:2013} dataset. We follow the evaluation protocol and cross-validate and report the video and frame mAP results on three splits. We edit and use the evaluation script from \cite{peng2016multi}. Our models are fine-tuned from Kinetics-400 weights and are trained for 10 epochs. Table \ref{tab:jhmdb_splits} shows the video mAP scores of three of our models on JHMDB and demonstrates that the proposed ACAM model achieves the best performance.
% for all and the average across the three splits for video-mAP. 
Proposed ACAM model achieves the best performance across all implementations which is consistent with the AVA dataset results.

\begin{table}[b]
\begin{footnotesize}
\begin{center}
  \begin{tabular}{| l || c | c | c || c | }
    \hline
    JHMDB - Models      &  Split1  & Split2 & Split3 & AVG \\ \hline \hline
    I3D Head + RoIPool        & 77.57 &  73.91 & 75.64 & 75.71 \\ \hline
    I3D Head + RoIPool + Tail  & 80.53 & 81.44 & 80.77 & 80.91 \\ \hline
    \textbf{I3D Head + ACAM + Tail} & \textbf{84.68} & \textbf{83.78} & \textbf{83.30} & \textbf{83.92}  \\ \hline

  \end{tabular}
\end{center}
\caption{Video mAP results on 3 splits of JHMDB and the average.}
\label{tab:jhmdb_splits}
\end{footnotesize}
\end{table}

Table \ref{tab:jhmdb_sota} compares video and frame mAP of our ACAM model with state-of-the-art models. ACAM outperforms the by 3.80 video mAP and 1 frame mAP without optical flow.

\begin{table}[b]
\begin{small}
\begin{center}
  \begin{tabular}{| l || c |  c  |}
    \hline
    JHMDB - Models   & Frame mAP   & Video mAP \\ \hline \hline
    Action-RCNN\cite{peng2016multi}         & 58.5    & 73.1 \\ \hline
    ACT-Tubelet\cite{kalogeiton2017action}  & 65.7            & 73.7 \\ \hline
    I3D-RoI\cite{gu2018ava}     & 73.3        & 78.6 \\ \hline
    ACRN\cite{sun2018actor}     & 77.9        & 80.1 \\ \hline\hline
    \textbf{ACAM}       & \textbf{78.9}      & \textbf{83.92} \\ \hline

  \end{tabular}
\end{center}
\caption{ mAP values averaged across 3 splits of JHMDB dataset. }
\label{tab:jhmdb_sota}
\end{small}
\end{table}

\subsection{Real-Time Framework for Action Detection}\label{sec:pipeline}

\begin{figure}[t]
\begin{center}
\includegraphics[width=1.0\linewidth]{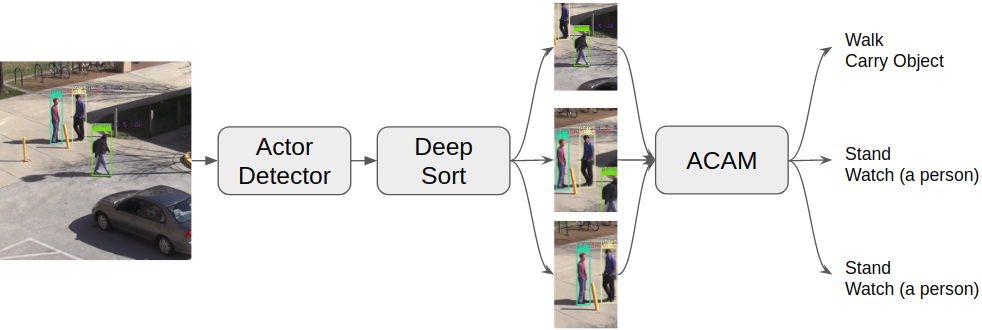}
\end{center}
\vspace*{-.25cm}
   \caption{ACAM action detection framework running on a surveillance video from VIRAT \cite{oh2011large}. Actors are detected by object detectors and tracked over frames by Deep Sort \cite{Wojke2017simple}. The generated tubes for each person is analyzed by the ACAM action detector.
   \vspace{-0.4cm}}
\label{fig:acam_framework}
\end{figure} 

We evaluate the transferability and performance of the proposed model on different datasets qualitatively. We implement an end-to-end framework for detecting and tracking actors and analyzing their actions. The action model in this section is trained on AVA dataset and is not fine-tuned on the other datasets which demonstrates transferability.

We combine the person detector with the Deep Sort \cite{Wojke2017simple} tracker. Deep Sort is a simple tracking/re-identifying model that uses a deep association metric for matching detected person bounding boxes. This allows us to track the detections over time and generates person tubelets. 

Since the proposed model explicitly models the surrounding context, a larger area than the person's tubelet is essential to model interactions. Due to the large view of surveillance videos, it is not feasible to process the entire scene. For this reason, square regions centered on the person's location and twice the size of the person's area are cropped and fed to the action detection framework.

The overall pipeline is shown in Fig. \ref{fig:acam_framework}. First, we extract the actor tubes with a larger context area from the video using the detector and the tracker. Then, each detected tube is analyzed by the ACAM module for actions. Input frame and cropped tubes for each actor are visualized from the VIRAT \cite{oh2011large} surveillance dataset. 
Notice that in interaction cases such as ``watching a person" the model benefits from having a person in the surrounding context.

\begin{figure}[t]
\begin{center}
\includegraphics[width=1.0\linewidth]{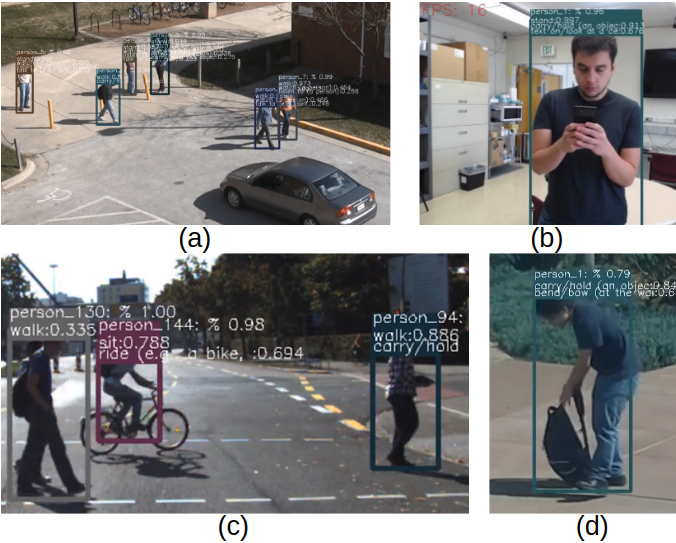}
\end{center}
\vspace*{-.25cm}
   \caption{Qualitative results of ACAM video action detection framework visualized on different sources. a) VIRAT surveillance dataset \cite{oh2011large}, b) Webcam inputs at 16 fps, c) KITTI \cite{Geiger2012CVPR} autonomous driving dataset, d) Campus Surveillance videos.}
\label{fig:framework_results}
\end{figure}

% \newpage
We provide additional qualitative results on Fig. \ref{fig:framework_results} for the autonomous driving dataset KITTI \cite{Geiger2012CVPR}, surveillance dataset VIRAT, webcam videos and campus surveillance. These videos sources are unseen for the action model which was trained on AVA (movies) dataset. Accurate qualitative results demonstrate the transferability of our framework. Videos are available in the supplementary material.

A real-time version of this pipeline is open-sourced and available at Demo Github. It achieves 16 frames per second through a webcam on a single Nvidia GTX 1080Ti GPU using a fast SSD \cite{liu2016ssd}-Mobilenet2 \cite{sandler2018mobilenetv2} object detector. This further demonstrates the advantage of modularity as the object detector can easily be changed for faster performance.

\subsection{Ablation Analysis} \label{sec:ablation}

\noindent \textbf{Actor Detection Performance:}  To test the performance of the actor detector, we calculate the detection AP of every actor for every class on validation set. 
% Averaging the detection result across classes instead of across number of samples allows us to accurately evaluate the detection rate for this task as there is a large variation in number of samples of each class. 
Table \ref{tab:object_detector_results} shows the detection frame AP scores for the AVA v2.1 validation set.

\begin{table}[htb]
\begin{footnotesize}
\begin{center}
  \begin{tabular}{| l || c  || c |}
    \hline
    Object Detector     & AVA Actor Detection AP & Speed(ms/frame)\\ \hline \hline
    F RCNN-NAS              & 97.10 & 1833 \\ \hline
    F RCNN-Resnet101        & 95.97 & 106 \\ \hline
    SSD - MobileNetV2            & 66.16 & 31\\ \hline

  \end{tabular}
\end{center}
\caption{ AP results for actor detection rate for different detectors and their detection speed. This demonstrates that detectors work well without fine tuning and shows the speed trade-off.}
\label{tab:object_detector_results}
\end{footnotesize}
\end{table}

\noindent \textbf{Transferable Actor Detection: } The main reason of using a pre-trained frozen person detector instead of training an RPN on the action dataset is transferability. During training, object detectors see large variations in objects from large datasets such as MS-COCO \cite{lin2014microsoft} compared to action datasets. This makes object detectors more transferable to other datasets compared to retrained RPNs. To test this hypothesis, we compare the actor detection rates of same model architecture with and without fine-tuning. AVA model is fine-tuned on top of the COCO weights whereas COCO model is frozen and is not fine-tuned. Table \ref{tab:object_detector_generalizable} shows their comparisons on different datasets. Even though fine-tuned model achieve slightly better actor detection rate on the AVA dataset, the performance degradation is significant on datasets such as VIRAT \cite{oh2011large} and KITTI \cite{Geiger2012CVPR}.

\begin{table}[htb]
\begin{footnotesize}
\begin{center}
  \begin{tabular}{| l || c |  c || c | }
    \hline
    Actor Detection    & F RCNN AVA  & F RCNN COCO  & $\Delta$\\ \hline \hline
    AVA               & 98.60 & 95.97  & $+2.63$ \\ \hline
    VIRAT           &  9.94 & 30.44 & $-20.50$\\ \hline
    KITTI            & 27.04 & 54.57 & $-27.53$\\ \hline
    
  \end{tabular}
\end{center}
\vspace*{-.25cm}

\caption{ AP results for actor detection rate of the same object detector trained on AVA and COCO and tested on different datasets. Actor detectors lose transferability to different domains when fine-tuned on action datasets (AVA in this case), which is shown by the difference ($\Delta$: F RCNN AVA - F RCNN COCO).}
\label{tab:object_detector_generalizable}
\end{footnotesize}
\end{table}

% \noindent \textbf{Modularity of Actor Detection: } By using a separate object detector and augmenting bounding box coordinates during training, we reduce the model's dependency on the specific object detector. This allows us to replace the object detector with a faster one during test time with minor performance hit. Table \ref{tab:object_detector_modular} shows the action detection performance on AVA v2.1 with different object detectors during test time. We can see that with some performance hit, we can switch the object detector for faster ones.

% \begin{table}[htb]
% \begin{footnotesize}
% \begin{center}
%   \begin{tabular}{| l || c |  c | c |}
%     \hline
%     Object Detector     & F RCNN NAS & F RCNN RESNET & SSD Incp\\ \hline
%     AVA Validation      &  22.67     & 21.90       & 11.00 \\ \hline

%   \end{tabular}
% \end{center}
% \caption{ ACAM model performance on the AVA v2.1 validation set with different object detectors during test time.}
% \label{tab:object_detector_modular}
% \end{footnotesize}
% \end{table}

\begin{figure}[htb]
\begin{center}
\includegraphics[width=1.0\linewidth]{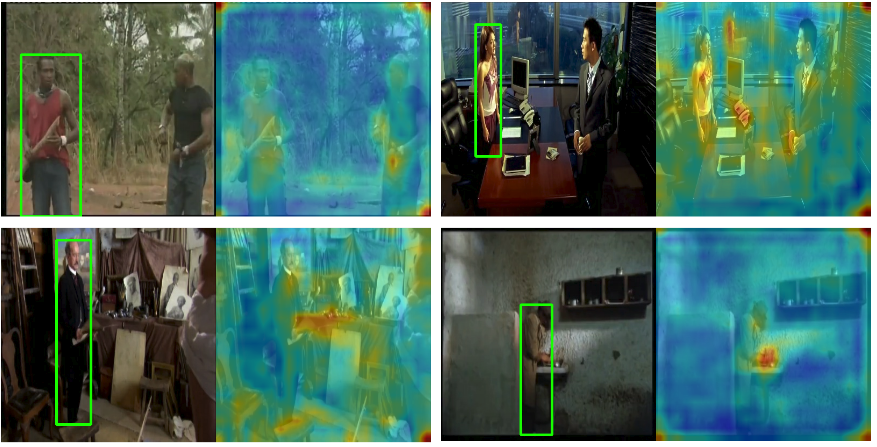}
\end{center}
\vspace*{-.25cm}
   \caption{Generated Actor Conditioned Attention Maps. Higher attention values are usually observed around objects (paper, chairs, teapot, phones), on faces and hands of the actors.}
\label{fig:attn_maps}
\end{figure}

\noindent \textbf{Visualization of Attention Maps: } In ACAM, an attention map for each feature channel is generated. This allows us to model different types of interactions efficiently. Since the feature maps are sparse, to visualize them, we average the attention map values across the feature dimension where they have non-zero values in their respective feature map. This generates a representation where each actor's relation with the scene is visible. Fig. \ref{fig:attn_maps} shows this visualization on different examples. Note that a higher attention value is obtained on objects and actor faces/hands.

\begin{figure}[t]
\begin{center}
\includegraphics[width=0.9\linewidth]{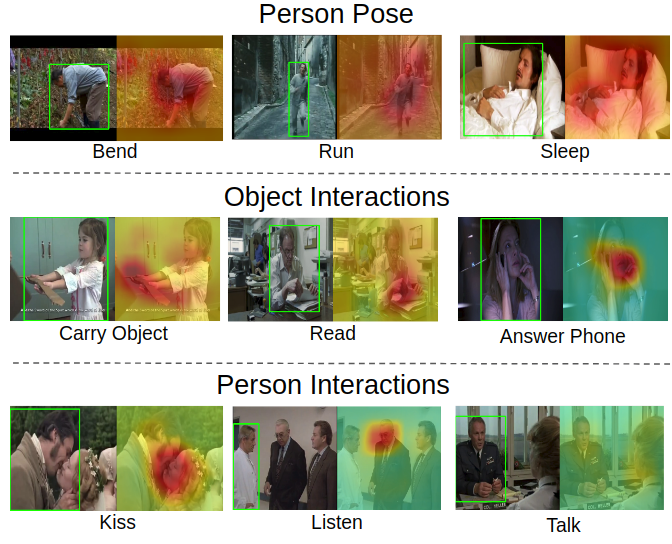}
\end{center}
\vspace*{-.25cm}
   \caption{Class activation maps for detected actors in validation set. Each image represents the activation maps for the actor annotated by the green box and the given class. Red regions on the activation maps represent larger values.
%   \vspace{-0.5cm}
}
\label{fig:cams_per_class}
\end{figure}

\noindent \textbf{Class Activation Maps: } Using the global pooling layer at the last layer, we can generate class activation maps for each class (similar to \cite{zhou2016learning}). We demonstrate activation maps for several different cases. Fig. \ref{fig:cams_per_class} shows activation maps for different categories of actions. Activation maps are shown for actors annotated in green bounding boxes. Maximum activations across timesteps are visualized in the figures as these activations are also time sequences.

We observe that pose actions such as run/bend get activated around the actor while object interaction actions such as carry object/read are activated around the relevant objects and the actors. Person interactions also show some interesting results. The passive actions such as ``watch a person", ``listen to a person" gets activated where there is another person in the scene that is ``talking" or relevant.

\begin{figure}[t]
\begin{center}
\includegraphics[width=1.0\linewidth]{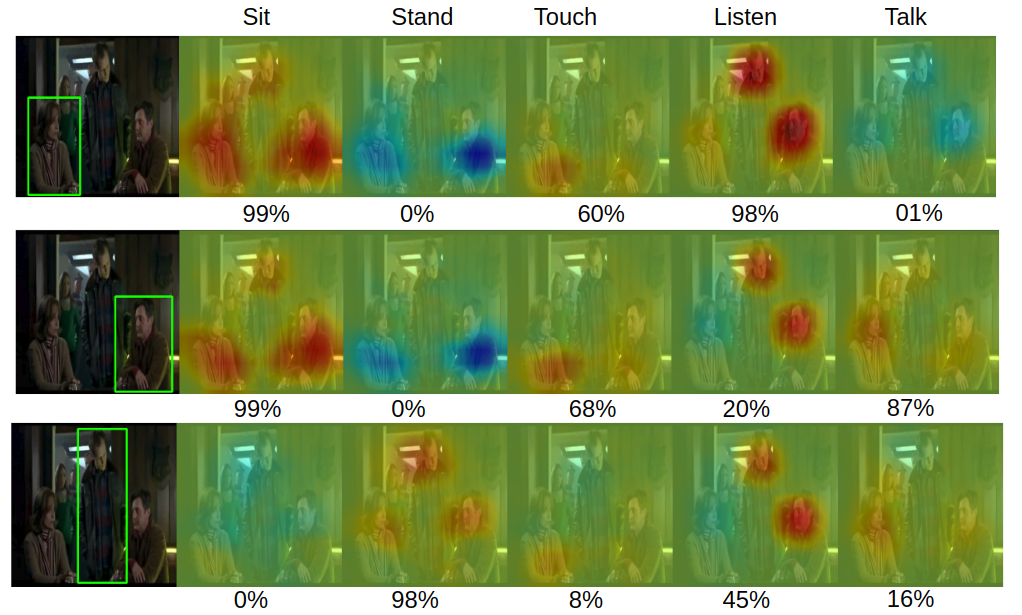}
\end{center}
\vspace*{-.25cm}
   \caption{Class activation maps for detected actors. Each row represents the activation maps for the actor annotated by the green bounding box. Person on the right is talking in the video. Red regions on the activation maps represent the higher values.
%   \vspace{-0.5cm}
   }
\label{fig:cams_per_actor}
\end{figure}

Fig. \ref{fig:cams_per_actor} shows a scene with three people and their conditioned activation maps for specific actions. Each row represents the activation maps that are conditioned on the person in the green bounding box. We observe that complementary actions such as ``talking'' and ``listening'' gets activated on the person with the opposite action. This is due to our model architecture. As we initially extract the actor feature vector $\mathbf{r}_a$ from the actor's location, this feature vector contains the information that the current actor $a$ is ``listening''. Therefore the attention map generated from actor's vector $\mathbf{r}_a$ and context $\mathbf{E}$ looks for a person that is ``talking'' and focuses the attention on those locations.

% This is also visible in Figure \ref{fig:cams_per_actor}. Model is correctly able to detect that the person on the right is talking, this allows the ``listening" action on other actors to get activated due to the presence of this talking person. 

% \textbf{Failure Cases: }

% \textbf{Qualitative results on different datasets/videos: } We qualitatively test our model on different datasets to observe the generalizibility to different video sources. We combine our object detector with a basic tracker and test the action detection. Figure \ref{} shows results on autonomous driving dataset KITTI and surveillance dataset VIRAT. 

% \noindent \textbf{Single attention map vs per feature channel attention map: } Works modeling contextual interactions such as \cite{wang2018non} use a single function to model interactions between two points. However, this prevents the model from detecting different types of interactions. In order to address this issue, instead of generating a single attention map in our work, we generate the attention maps per feature channel. When trained with a single attention map that multiplies the input feature map $I$, the model was unable to distinguish between different actors in the scene and generated same confidence values for every actor. 

\section{Discussion}
\subsection{ Comparisons of Attention Mechanisms }

We compare ACAM with similar attention studies including: Actor-Centric Relation Network (ACRN) \cite{sun2018actor}, Attentive Contexts for Object Detection (ACOD) \cite{li2017attentive} and Non-Local Neural Networks (NL) \cite{wang2018non}. 
% In this section, we analyze the differences of these works to our proposed model and address their shortcomings for this specific task of action detection. 

\textbf{ACRN} uses the Relation module (Eq. \ref{eq:relation}) without the sigmoid function to generate relation features for actors and context. The relation features are directly used for action classification with convolutional layers which are trained from scratch. In contrast, ACAM leverages the relation features to generate attention maps, which amplify/dampen the I3D feature map $\mathbf{I}$. The actor conditioned features $\mathbf{F}$ are scaled versions of $\mathbf{I}$ and the model is able to effectively leverage the pre-trained I3D-Tail on $\mathbf{F}$.
% \textbf{ACRN} uses a similar relation structure for contextual information. It combines the actor and contextual features and trains an additional convolutional layer for classification. In contrast, we combine actor and contextual features to generate a set of weights for the original features. These weights characterize the actor-action relationships while preserving the features. This allows us to use the remaining convolutional layers from the CNN back-bone.% while conditioning it on the locations relevant to the actor. 

\textbf{ACOD} uses an LSTM branch to generate an attention map. It leverages context to generate a single global feature for every region proposal. This ``attention branch" omits the individual differences of region proposals. ACAM generates attention maps by conditioning the context on each detected region (actor) individually to represent individual interactions which is better suited for action detection tasks.

% \textbf{NL} tackles action recognition using weighted means of every location in the feature map. This captures the relevance between every pixel pair. In contrast, the proposed ACAM represents the actor-context interactions in a higher level and presents this information as attention maps.
% Using and vectorizing features from actor's bounding box is a more efficient representation than looking for every pixel location on the sparse actor feature map. Using weighted sums and softmax functions to generate the weights limits the number of available interactions. ACAMS deals with different relations by using a sigmoid function for attention.
    
\textbf{NL}, similar to the self-attention \cite{vaswani2017attention}, uses matrix multiplication to find relations among pixel pairs in a spatio-temporal feature tensor. The relations are normalized through a softmax function to emulate an attention map. In an action detection setting where the actions are focused on actors, instead of finding relations among all pixel pairs, it is more effective to find relations between all context pixels and condition on individual actor features as in ACAM.

% This limits the number of available interactions as different relations will be competing with each other, in contrast to a sigmoid.

\subsection{Summary}
We presented a novel action detection model that explicitly captures the contextual information of actor surroundings. The proposed ACAM uses attention maps as a set of weights to highlight the spatio-temporal regions that are relevant to the actor, while damping irrelevant ones. This method is presented as a replacement to RoIPooling. ACAM  is more suited for preserving interactions with surrounding context such as objects, other actors and scene. We demonstrated through thorough experimentation that ACAM improves the performance on multiple datasets and outperforms the state-of-the-art. We implemented and open-sourced a real-time atomic action detection pipeline to demonstrate the feasibility and modularity of ACAM. 

\section{Summary of Variables}

Following table summarizes the variables used in the proposed method and their shapes for our implementation. These variables and their shapes are explained clearly in the paper and this table helps the reader to follow the proposed methodology.

\begin{table}[!h]
\begin{footnotesize}
\begin{center}
  \begin{tabular}{| c | c | c | }
    \hline
    \textbf{Symbol} & \textbf{Definition} & \textbf{Shape} \\ \hline
    \hline
    $ V $   & Input Video & $ 32 \times 400 \times 400\times 3$ \\ \hline 
    $ a $   & Actor  & $ -$ \\ \hline 
    $ \textbf{I} $  & I3D Feature Map &  $8\times25\times25\times832$ \\ \hline
    \hline
    $ RoI(\textbf{I},a) $  & RoI Features on Actor &  $1\times10\times10\times832$ \\ \hline
    $ \textbf{w}_\rho $  & Weights for actor &  $83200\times208$ \\ \hline
    $ \textbf{b}_\rho $  & Biases for actor &  $208$ \\ \hline
    $ \textbf{r}_a $  & Actor Feature vector &  $208$ \\ \hline
    \hline
    $ \textbf{w}_\eta $  & Weights for Context &  $1\times1\times1\times832\times208$ \\ \hline
    
    $ \textbf{b}_\eta $  & Biases for Context &  $1\times1\times1\times208$ \\ \hline
    $ \textbf{E} $  & Context Feature Tensor &  $8\times25\times25\times208$ \\ \hline 
    \hline
    $ \textbf{w}_\Omega $  & Actor weights for Relation &  $1\times1\times1\times208\times832$ \\ \hline
    
    $ \textbf{w}_\gamma $  & Context weights for Relation &  $1\times1\times1\times208\times832$ \\ \hline
    
    $ \textbf{b}_\beta $  & Biases for Relation &  $1\times1\times1\times832$ \\ \hline
    
    $ \textbf{R}_a $  & Relation Features &  $8\times25\times25\times832$ \\ \hline
    
    $ \textbf{ACAM}_a $  & Attention Maps for Actor $a$ &  $8\times25\times25\times832$ \\ \hline
    
    $ \textbf{F}_|a $  & Actor Conditioned Features &  $8\times25\times25\times832$ \\ \hline
    \hline
    
    $ T $  & Temporal Resolution of $\textbf{I}$ &  $8$ \\ 
           &  indexed by $t$           & \\\hline
           
    $ H $  & Spatial Height of $\textbf{I}$ &  $25$ \\ 
           &  indexed by $h$           & \\\hline
    $ W $  & Spatial Width of $\textbf{I}$ &  $25$ \\ 
           &  indexed by $w$           & \\\hline
    \hline 
    
    $ C $  & Feature Channel dimension of $\textbf{I}$ &  $832$ \\ \hline
    
    $ N $  & Feature Channel dimension of $\textbf{r}_a$ &  $208$ \\ 
           &  set to be $C/4$           & \\\hline
    $ M $  & Feature Channel dimension of $\textbf{E}$ &  $208$ \\ 
           &  set to be $C/4$           & \\\hline
           
    % \hline

  \end{tabular}
\end{center}
\caption{Table of variables used in proposed model and their respective shapes used in our implementation. }
\label{tab:published_map_results}
\end{footnotesize}
\end{table}

{\small
\bibliographystyle{ieee}
\bibliography{output}
}

\end{document}